\documentclass[10pt,a4paper,logo]{googledeepmind}

\usepackage{microtype}
\ifPDFTeX\else
  \microtypesetup{tracking=false}
\fi
\usepackage{latexsym}
\usepackage{multirow}
\usepackage{float}
\usepackage{algorithm}
\usepackage{algorithmic}
\usepackage{natbib}
\usepackage{adjustbox}
\usepackage{placeins}
\usepackage{wrapfig}
\usepackage{needspace}
\definecolor{mygreen}{RGB}{0,128,0}
\definecolor{myred}{RGB}{180,0,0}
\definecolor{mygray}{gray}{0.92}
\definecolor{visualtok}{RGB}{237,214,230}
\definecolor{reasontok}{RGB}{216,232,255}
\definecolor{anchortok}{gray}{0.92}
\newcommand{\inc}[1]{\textcolor{mygreen}{\scriptsize{$\uparrow$#1}}}
\newcommand{\dec}[1]{\textcolor{myred}{\scriptsize{$\downarrow$#1}}}
\newcommand{\vistok}[1]{\begingroup\setlength{\fboxsep}{1pt}\colorbox{visualtok}{\strut #1}\endgroup}
\newcommand{\reatok}[1]{\begingroup\setlength{\fboxsep}{1pt}\colorbox{reasontok}{\strut #1}\endgroup}
\newcommand{\anchortoken}[1]{\begingroup\setlength{\fboxsep}{1pt}\colorbox{anchortok}{\strut #1}\endgroup}

\graphicspath{{images/}{figures/}}
\hypersetup{
  pdftitle={Token-Disentangled Latent Test-Time Scaling for Vision-Language Reasoning},
  pdfauthor={Hao-Xuan Ma, Yihao Liu, Yutao Sun, Yanting Miao, Mengyu Zhou, YiCheng Xiao, Long Chen, Zhenguo Li, Han-Jia Ye, Xiaoxi Jiang, Guanjun Jiang}
}

\title{Token-Disentangled Latent Test-Time Scaling for Vision-Language Reasoning}

\author[1,2,3]{Hao-Xuan Ma\textsuperscript{\ensuremath{\ddagger}}}
\author[1]{Yihao Liu\textsuperscript{\ensuremath{\ddagger}}}
\author[1,4]{Yutao Sun\textsuperscript{\ensuremath{\ddagger}}}
\author[1,5]{Yanting Miao\textsuperscript{\ensuremath{\ddagger}}}
\author[1]{Mengyu Zhou\textsuperscript{\ensuremath{\dagger}}}
\author[6]{YiCheng Xiao}
\author[7]{Long Chen}
\author[7,8]{Zhenguo Li}
\author[2,3]{Han-Jia Ye\textsuperscript{\ensuremath{\dagger}}}
\author[1]{Xiaoxi Jiang}
\author[1]{Guanjun Jiang}
\affil[1]{Qwen Business Unit of Alibaba}
\affil[2]{School of Artificial Intelligence, Nanjing University}
\affil[3]{National Key Laboratory for Novel Software Technology, Nanjing University}
\affil[4]{Zhejiang University}
\affil[5]{University of Waterloo}
\affil[6]{Chinese Academy of Sciences}
\affil[7]{The Hong Kong University of Science and Technology}
\affil[8]{Frontier Robotics}
\footnotetext{\textsuperscript{\ensuremath{\ddagger}}Core contributors.\quad
  \textsuperscript{\ensuremath{\dagger}}Corresponding authors.}
\correspondingauthor{\texttt{yehj@lamda.nju.edu.cn}; \texttt{zhoumengyu.zmy@alibaba-inc.com}}

\begin{abstract}
  Latent test-time scaling improves reasoning by refining hidden states during inference, but existing methods typically apply a single scalar reward to all editable latent tokens. 
  For multimodal large language models, this global update ignores that generated tokens play different roles: some are sensitive to visual evidence, while others correspond to uncertain reasoning decisions. We present \textbf{Token-Disentangled Latent Test-Time Scaling}, an inference-time framework that makes latent refinement token-role-aware. 
  Starting from an initial generated trajectory, we optimize a short hidden-state prefix while routing perception-side visual feedback to image-sensitive tokens and reasoning feedback to high-entropy tokens. Tokens selected by neither route are constrained by an anchor regularizer. 
  Across both perception and reasoning benchmarks on Qwen2.5-VL-7B and InternVL3.5-8B, our method lifts macro accuracy over CoT by $+2.57$ and $+1.51$ respectively, and outperforms strong output-space test-time scaling baselines under matched decoded-candidate budgets.
  Code is available at \url{https://github.com/Qwen-Applications/TD-LTTS}.
\end{abstract}

\begin{document}
\maketitle
\suppressfloats[t]
\begin{figure*}[t]
    \centering
    \includegraphics[width=\textwidth]{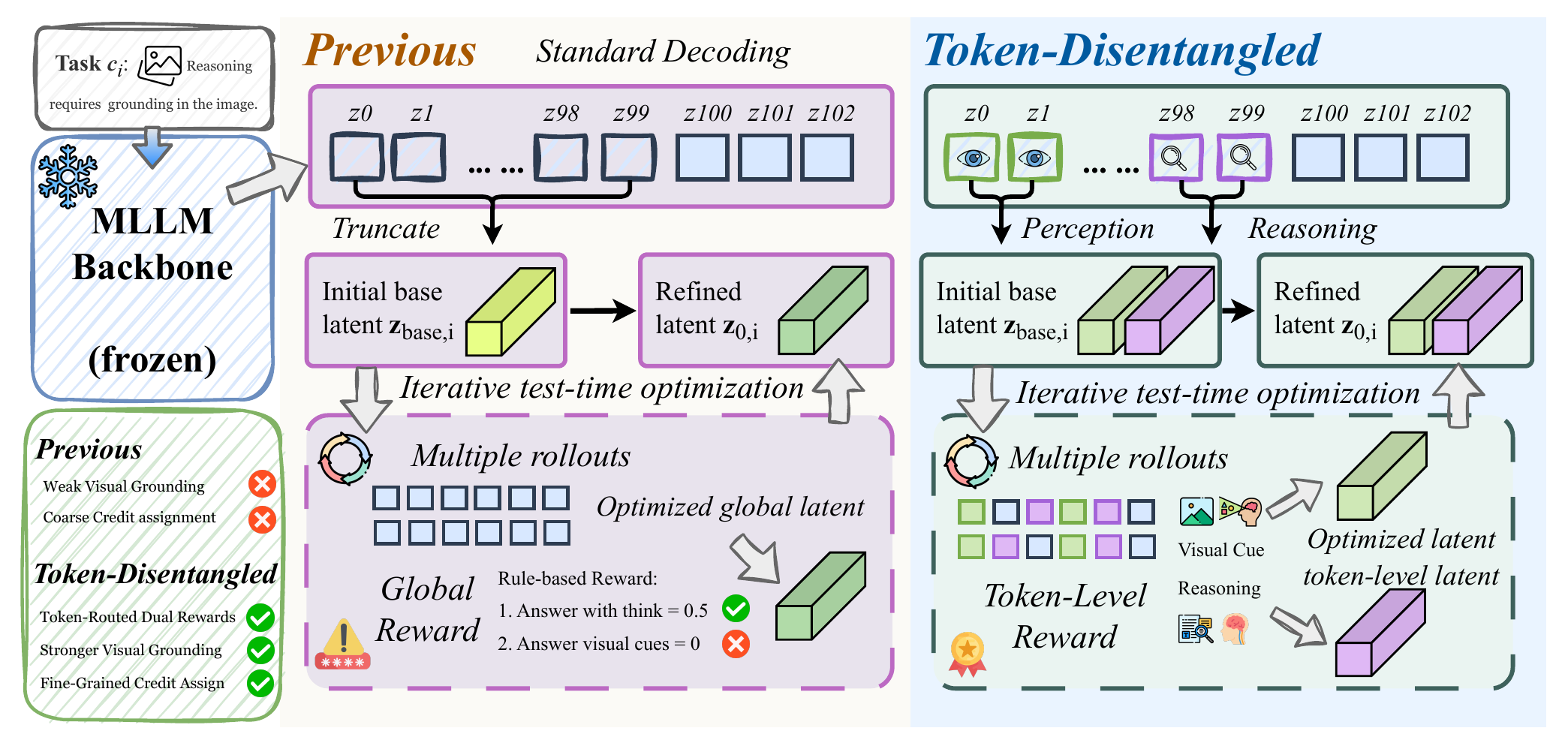}
    \caption{\textbf{From global latent refinement to token-disentangled latent refinement.} Our method augments latent test-time scaling with token-role-aware routing, separate reasoning and visual rewards, and anchor regularization, improving visual engagement, answer discrimination, and consistency across multimodal reasoning benchmarks.}
    \label{fig:overview}
\end{figure*}

\section{Introduction}

Multimodal large language models (MLLMs) have recently achieved impressive progress in integrating visual and textual information for complex reasoning tasks~\citep{shao2024visual,deng2025openvlthinker,wen2025reinforcement}.
Despite this progress, further improving these capabilities through additional training remains costly, motivating test-time scaling methods that keep the model frozen while allocating extra computation to the same input at inference time~\citep{brown2024large,geiping2026scaling}.
In multimodal reasoning, however, extra inference-time computation is useful only when it targets the source of the model's error~\citep{liu2025reasoning}. Such errors can arise from distinct bottlenecks: a model may generate a coherent answer while relying on insufficient visual evidence, or it may capture the relevant visual cues but still fail to discriminate among competing answers.

Recent latent test-time scaling methods move inference-time computation into hidden states, refining internal representations rather than updating model parameters~\citep{li2025seek,zhang2025latentevolve}.
This offers a natural way to reuse existing MLLMs, since the search operates over instance-specific hidden states while leaving the model frozen.
However, existing latent refinement objectives typically optimize a short sequence of editable latent tokens with a single global scalar reward, as illustrated in Figure~\ref{fig:overview}.
This global feedback is too coarse for multimodal reasoning, where different tokens can play different roles: some are closely tied to visual evidence, while others support reasoning and answer discrimination~\citep{lu2026bridging,huang2025spotlight}.
As a result, tokens with different roles may receive the same credit signal even when they contribute to the error in different ways.
More concretely, a candidate-level reward cannot determine whether a failed continuation stems from insufficient visual grounding or flawed answer discrimination, which can assign refinement pressure to latent positions unrelated to the actual failure mode.

These observations suggest that effective latent test-time scaling for MLLMs should refine both visual-evidence use and reasoning states.
To this end, we propose \textbf{Token-Disentangled Latent Test-Time Scaling}, an inference-time framework that makes latent refinement token-role-aware.
Our method retains the standard latent test-time optimization: an MLLM first generates an initial response, a short slice of generated hidden states is selected as editable latent variables, and this slice is refined by decoding and scoring candidate continuations.
Instead of broadcasting one global reward to all editable tokens, we decompose the optimization signal into visual-evidence and reasoning components and route them to different latent positions.
Visual tokens are selected by their sensitivity to image perturbations, while reasoning tokens are selected by output uncertainty.
This enables visual and reasoning feedback to update different parts of the latent sequence.

We evaluate our method on six multimodal reasoning benchmarks spanning both perception-heavy and reasoning-heavy settings, using Qwen2.5-VL-7B and InternVL3.5-8B as frozen backbones.
Across these benchmarks, our method consistently improves the initial CoT rollout, yielding macro-average gains of +2.57 on Qwen2.5-VL-7B and +1.51 on InternVL3.5-8B.
Under the same decoded-candidate budget, our method also outperforms strong output-space test-time scaling baselines on both backbones, suggesting that targeted latent refinement can improve the model's internal computation rather than merely selecting among sampled answers.
Further analyses show that both visual-token routing and reasoning-token routing contribute to the final improvement.

Our contributions are threefold:
\begin{itemize}
    \item We identify a multimodal limitation of existing latent test-time scaling: candidate-level scalar rewards cannot distinguish between visual-evidence and reasoning roles within the editable latent sequence.
    \item We introduce token-disentangled latent refinement, which routes visual engagement feedback to image-sensitive tokens and reasoning feedback to uncertain reasoning tokens.
    \item We demonstrate consistent gains over CoT across six multimodal reasoning benchmarks and two frozen MLLM backbones, showing the effectiveness of role-aware latent refinement.
\end{itemize}

\section{Related Work}

\subsection{Latent Test-Time Scaling}
Test-time scaling improves reasoning by spending extra computation during inference rather than increasing model size. Early work mainly scaled in output space through repeated sampling, search, or self-correction~\citep{wang2022self,brown2024large,welleck2022generating}, with later studies analyzing compute-optimal allocation~\citep{snell2024scaling}. A parallel line moves this computation into continuous latent states~\citep{geiping2026scaling,you2025parallel,muennighoff2025s1}: Coconut~\citep{hao2024training} uses continuous thoughts as reusable reasoning states, while SoftCoT and SoftCoT++ introduce soft latent reasoning and test-time latent exploration~\citep{xu2025softcot++,xu2025softcot}. More recent methods refine latent states through search or policy optimization, including LatentSeek~\citep{li2025seek}; multimodal variants such as DMLR~\citep{liu2025reasoning} and VaLR~\citep{jeon2026vision} further emphasize preserving visual information during latent computation~\citep{ahmadpour2025limits,zhang2025alphaone,kaya2025efficient}. These studies show the promise of latent refinement, but most apply sequence-level, step-level, or globally coupled objectives. Our work instead routes perception and reasoning feedback to different latent tokens.

\subsection{Token-Level Credit Assignment in Multimodal Reasoning}

Reliable multimodal reasoning requires more than producing a plausible final answer: intermediate decisions must remain grounded in the relevant visual evidence. Prior studies show that stronger language reasoning does not eliminate perception bottlenecks, and that only a subset of visual tokens is often decisive for downstream prediction~\citep{xiong2025llava,tong2024eyes,jiang2025kind}. This issue becomes more pronounced in generated reasoning trajectories, where longer chains can dilute visual attention and amplify hallucination~\citep{liu2025visual}.

These findings motivate token-level credit assignment for multimodal reasoning. Perception-aware optimization and token analyses identify sparse but pivotal visual dependencies~\citep{huang2025spotlight}, while ToR~\citep{lu2026bridging} and PRCO~\citep{miao2026seeing} explicitly separate perception- and reasoning-related roles during training. Our work brings this role distinction into latent test-time scaling: instead of assigning one candidate-level reward to all editable hidden states, we route perception-side and reasoning-side feedback to different latent tokens of a frozen MLLM during inference.

\section{Preliminaries}
\label{sec:prelim}

\noindent\textbf{Problem Setup.}
We first formulate latent test-time scaling in the conventional text-only setting.
Given a textual context $c$, a language model with fixed parameters $\theta$
generates a response $y=(y_1,\ldots,y_T)$ with hidden states
$h_1,\ldots,h_T$. At each position, the LM head maps the hidden state to
the next-token distribution $p_\theta(y_t\mid c,y_{<t})$.

Latent test-time scaling treats a short prefix of these pre-LM-head hidden
states as editable latent variables,
\begin{equation}
    z^{(0)}=[h_1,\ldots,h_L],
    \qquad
    z^{(k)}=\{z_i^{(k)}\}_{i=1}^{L},
\end{equation}
where $L$ is the number of editable latent tokens and $k$ indexes refinement
steps. The model parameters remain frozen; only the instance-specific latent
prefix is updated.

Given a current latent prefix $z^{(k)}$, each editable vector induces a local
LM-head distribution
\begin{equation}
    \pi_i(v\mid z_i^{(k)})
    =
    \operatorname{softmax}(W_{\mathrm{lm}}z_i^{(k)})_v.
\end{equation}
Decoding from these distributions yields a latent-induced prefix
$\hat y_{1:L}$, after which the frozen model continues generation to produce a
candidate response $x$. We write the induced candidate distribution as
$p_\theta(x\mid z^{(k)},c)$ and the candidate set at step $k$ as
\begin{equation}
    \mathcal{X}^{(k)}=\{x^{(k)}_1,\ldots,x^{(k)}_M\}.
\end{equation}
A verifier or rule-based reward then assigns a scalar score $R(x,c)$ to each
decoded candidate. Prior latent search methods use this scalar to update the
entire latent prefix with an objective of the form
\begin{equation}
    \mathcal{L}
    =
    -R(x,c)
    \sum_{i=1}^{L}
    \log \pi_i(\hat y_i\mid z_i).
\end{equation}
This formulation is natural for unimodal reasoning: the reward is defined at
the candidate level, and all editable latent positions are optimized toward the
same decoded trajectory-level signal.

\noindent\textbf{Modality Gap in Vision-Language Models.}
For vision-language reasoning, the context becomes $\xi=(I,q)$ with image
$I$ and question $q$. The generated trajectory is now supported by mixed
evidence: some latent tokens are tied to visual input---their likelihoods
shift under image corruption, indicating that they encode objects,
attributes, counts, or spatial relations---while others correspond to
uncertain answer comparisons or reasoning transitions, identifiable by
high entropy in their LM-head distributions. A single scalar reward can
score answer quality or visual engagement, but it cannot tell which latent
positions should absorb perception feedback and which should absorb
reasoning feedback; broadcasting one signal to both groups risks mixing
the two updates and worsening the modality gap.

We therefore treat the latent prefix as a mixture of roles and perform
token-level credit assignment. We use two reward signals---a reasoning
reward $R_{\mathrm{rea}}$ for answer quality and a visual reward
$R_{\mathrm{vis}}$ for perception-side engagement---and route them to two
disjoint token groups: high-entropy tokens receive reasoning feedback,
while image-sensitive tokens receive visual feedback. This sets up the
token-disentangled latent update in Section~\ref{sec:method}.

\section{Method}
\label{sec:method}

We present \emph{Token-Disentangled Latent Test-Time Scaling}, an
inference-time adaptation method for vision-language reasoning. Building on
the latent test-time scaling formulation in Section~\ref{sec:prelim}, our
method changes how reward feedback is assigned to editable latent tokens.
Instead of applying one scalar reward to the entire latent prefix, we separate
visual feedback from reasoning feedback and route
each signal to the tokens most suited for it.

\begin{figure*}[t]
    \centering
    \includegraphics[width=\textwidth]{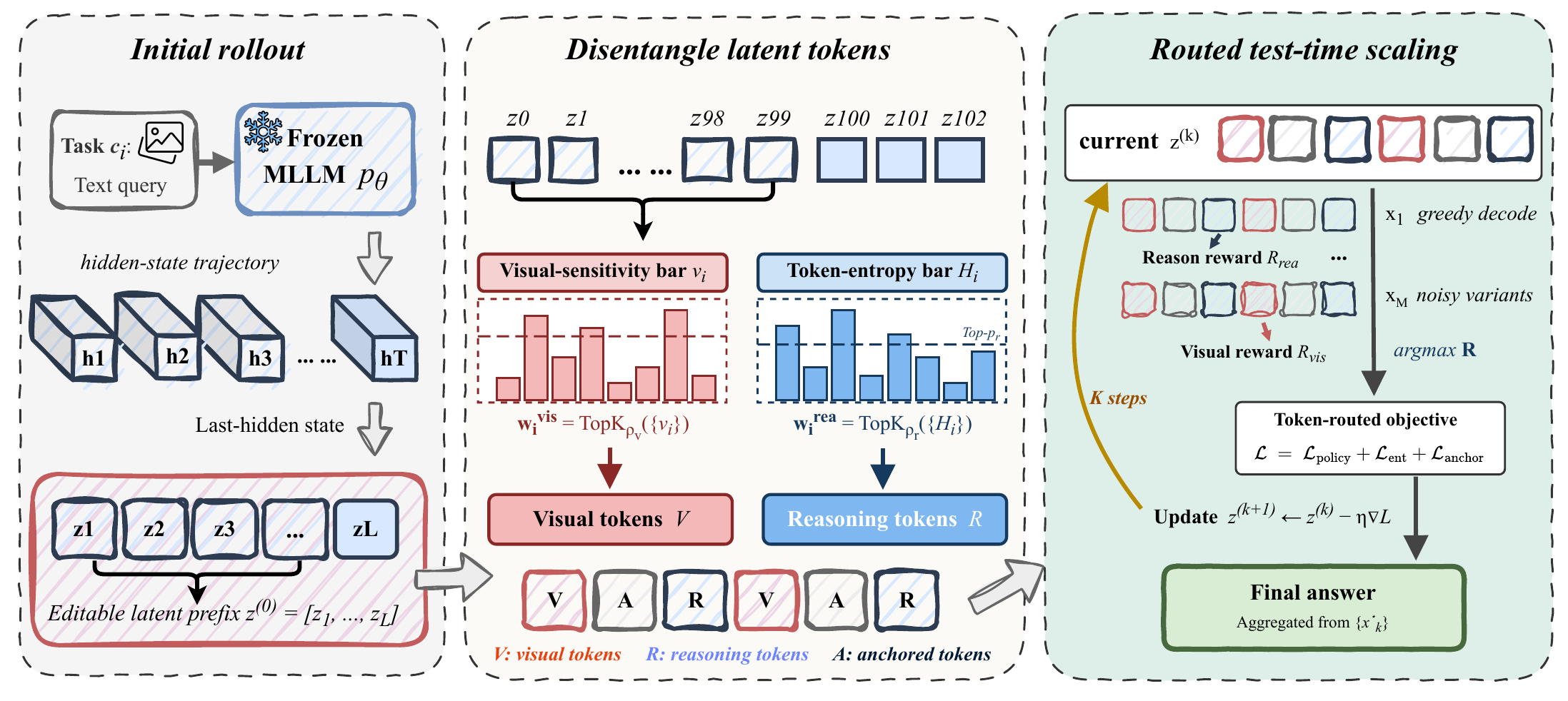}
    \caption{\textbf{Framework of Token-Disentangled Latent Test-Time Scaling.}
    A frozen VLM first produces an initial chain-of-thought response and its
    hidden-state trajectory. We refine an early latent prefix from this trajectory
    at test time. Visual tokens (identified by high image sensitivity) receive
    visual rewards, while reasoning tokens (identified by high entropy) receive
    reasoning rewards. Token routing prevents visual and
    reasoning feedback from
    collapsing into a single global latent update.}
    \label{fig:framework}
\end{figure*}

\subsection{Token-Routed Latent Refinement Objective}

The standard latent-search objective is to find an editable prefix
$z=(z_1,\ldots,z_L)$ of length $L$ that improves the expected quality of
decoded continuations:
\begin{equation}
    z^\star
    =
    \arg\max_z
    \mathbb{E}_{x\sim p_\theta(x\mid z,\xi)}
    [R(x,\xi)].
\end{equation}
Applying a single reward $R$ to every latent token, however, assumes that all
tokens in the slice play the same role. This assumption is especially weak for
multimodal reasoning: some latent tokens primarily carry visual evidence, while
others encode uncertain answer decisions or reasoning transitions.

We therefore decompose the context-conditioned feedback into a visual reward
and a reasoning reward:
\begin{equation}
    R_{\mathrm{vis}}(x;\xi,y),
    \qquad
    R_{\mathrm{rea}}(x;\xi),
\end{equation}
where $\xi=(I,q)$ is the multimodal context and $y$ is the initial rollout used
as the per-instance visual baseline. We write these rewards as
$R_{\mathrm{vis}}(x)$ and $R_{\mathrm{rea}}(x)$ when the context is clear,
and compute two disjoint token masks:
\begin{equation}
    w_i^{\mathrm{vis}}, w_i^{\mathrm{rea}} \in \{0,1\},
    \qquad
    w_i^{\mathrm{vis}}w_i^{\mathrm{rea}}=0.
\end{equation}
At update $k$, the feedback candidate is selected from a decoded
candidate set $\mathcal{X}^{(k)}$ by the reasoning reward:
\begin{equation}
    x_k^\star
    =
    \arg\max_{x\in\mathcal{X}^{(k)}}
    R_{\mathrm{rea}}(x).
\end{equation}
For $k\ge 1$, $\mathcal{X}^{(k)}$ is decoded from the current prefix
$z^{(k)}$ after the $k$-th latent update; for $k=0$ we set
$\mathcal{X}^{(0)}=\{y\}$, so the first update is driven by the rollout's
own reward without any extra decoding at $z^{(0)}$. The rollout $y$ is also
retained throughout as the per-instance reference for relative prediction
margins. The selected $x_k^\star$ is then used for both reward
channels: $R_{\mathrm{rea}}$ keeps answer quality as the selection
criterion, while $R_{\mathrm{vis}}$ measures whether that candidate
improves visual engagement relative to the initial rollout $y$.

During decoding, we also retain the editable-token prefix that produced each
candidate. Let $\hat y_{k,i}^{\star}$ denote the token at editable position $i$
for the selected candidate $x_k^\star$, with its log probability evaluated
under the current latent distribution $\pi_i(\cdot\mid z_i^{(k)})$. We first
define the routed log-probability sums
\begin{equation}
    \begin{aligned}
    S_{\mathrm{vis}}^{(k)}
    &=
    \sum_i
    w_i^{\mathrm{vis}}
    \log \pi_i(\hat y_{k,i}^{\star}\mid z_i^{(k)}), \\
    S_{\mathrm{rea}}^{(k)}
    &=
    \sum_i
    w_i^{\mathrm{rea}}
    \log \pi_i(\hat y_{k,i}^{\star}\mid z_i^{(k)}).
    \end{aligned}
\end{equation}
The token-routed policy loss is
\begin{equation}
    \begin{aligned}
    \mathcal{L}_{\mathrm{policy}}
    ={}&-
    \lambda_{\mathrm{pg}}^{\mathrm{vis}}
    R_{\mathrm{vis}}(x_k^\star)
    S_{\mathrm{vis}}^{(k)} \\
    &-
    \lambda_{\mathrm{pg}}^{\mathrm{rea}}
    R_{\mathrm{rea}}(x_k^\star)
    S_{\mathrm{rea}}^{(k)}.
    \end{aligned}
\end{equation}
Visual rewards therefore update image-sensitive tokens, while reasoning
rewards update high-entropy tokens.

We additionally regularize the update with a reasoning-weighted entropy term
and an anchor penalty:
\begin{equation}
    \mathcal{L}_{\mathrm{ent}}
    =
    \lambda_{\mathrm{ent}}
    \frac{
        \sum_i w_i^{\mathrm{rea}} H_i
    }{
        \sum_i w_i^{\mathrm{rea}}+\epsilon
    },
\end{equation}
\begin{equation}
    \mathcal{L}_{\mathrm{anchor}}
    =
    \lambda_{\mathrm{anchor}}
    \|z^{(k)}-z^{(0)}\|_2^2,
\end{equation}
where
\begin{equation}
    H_i
    =
    -
    \sum_v
    \pi_i(v\mid z_i^{(k)})
    \log \pi_i(v\mid z_i^{(k)}).
\end{equation}
The final objective minimized at each refinement step is
\begin{equation}
    \mathcal{L}
    =
    \mathcal{L}_{\mathrm{policy}}
    +
    \mathcal{L}_{\mathrm{ent}}
    +
    \mathcal{L}_{\mathrm{anchor}}.
\end{equation}

\subsection{Visual and Reasoning Rewards}

\paragraph{Visual reward.}
The visual reward uses image-token attention as a lightweight
visual-engagement proxy. Let $\mathcal{P}_I$ be the image-token positions. For a generated
trajectory $x=(u_1,\ldots,u_N)$, define
\begin{equation}
    e_j(x)
    =
    \frac{1}{M}
    \sum_{m=1}^{M}
    \sum_{p\in\mathcal{P}_I}
    A^{(\ell)}_{m,j,p}(I,x),
\end{equation}
where $A^{(\ell)}$ is the layer-$\ell$ attention tensor and $M$ is the number
of heads. With $N'=\min(N,N_{\max})$, the trajectory-level engagement score is
\begin{equation}
    E(x)
    =
    \frac{1}{N'}
    \sum_{j=1}^{N'}e_j(x).
\end{equation}
Using the initial rollout $y$ as the per-instance baseline, we define
\begin{equation}
    R_{\mathrm{vis}}(x)
    =
    \tanh\left(
        \frac{E(x)-E(y)}{\tau_{\mathrm{vis}}}
    \right).
\end{equation}
This relative reward also provides visual feedback for the initial
candidate set $\mathcal{X}^{(0)}$; baseline details are in
Appendix~\ref{sec:appendix-visual-baseline}. The bounded $\tanh$ caps the
update magnitude so a single instance with degenerate $E(y)$ cannot
dominate the latent step.

\paragraph{Reasoning reward.}
For a decoded candidate $x$, let $c(x)$ be its normalized short prediction.
We score candidate quality by subtracting a text-only prior from the
image-conditioned likelihood:
\begin{equation}
    S(c)
    =
    \log p_\theta(c\mid I,q)
    -
    \lambda_{\mathrm{p}}
    \log p_\theta(c\mid q).
\end{equation}
We combine a centered bounded score with a relative margin over competing
predictions:
\begin{equation}
    R_{\mathrm{rea}}(x)
    =
    \lambda_s\bar r(x)
    +
    \lambda_{\mathrm{mar}}r_{\mathrm{mar}}(c(x)).
\end{equation}
Here $\bar r(x)$ is derived from the score $S(c(x))$, while
$r_{\mathrm{mar}}$ compares $c(x)$ against alternatives from the same
refinement step. We give the full bounded-score and margin definitions in
Appendix~\ref{sec:appendix-reasoning-reward}.

\subsection{Token Routing}

\paragraph{Visual tokens.}
We identify visual tokens by measuring how sensitive their likelihoods are to
image corruption. Under the original image, the teacher-forced log probability
of the generated token corresponding to editable position $i$ is
\begin{equation}
    \ell_i
    =
    \log p_\theta(y_i\mid I,q,y_{<i}).
\end{equation}
After corrupting the image to obtain $\tilde I$, we recompute
\begin{equation}
    \tilde{\ell}_i
    =
    \log p_\theta(y_i\mid \tilde I,q,y_{<i}).
\end{equation}
The image-sensitivity score is
\begin{equation}
    v_i
    =
    \phi(\tilde{\ell}_i-\ell_i),
    \qquad
    \phi(x)=\exp(x)-x-1.
\end{equation}
Tokens with the highest image-sensitivity scores are routed to the visual
branch:
\begin{equation}
    w_i^{\mathrm{vis}}
    =
    \mathbf{1}
    \left[
        i\in
        \operatorname{TopK}_{\rho_v}
        \left(\{v_j\}_{j=1}^{L}\right)
    \right].
\end{equation}

\paragraph{Reasoning tokens.}
Reasoning tokens are identified by uncertainty under the current latent LM-head
distribution. We standardize the entropy values within the editable slice,
\begin{equation}
    \bar H_i
    =
    \frac{H_i-\mu_H}{\sigma_H+\epsilon},
\end{equation}
where $\mu_H$ and $\sigma_H$ are the mean and standard deviation of
$\{H_j\}_{j=1}^{L}$. Let $\mathcal{N}_{\mathrm{vis}}$ denote the non-visual
tokens, and let $\bar H_{\mathcal{N}_{\mathrm{vis}}}$ be the corresponding
standardized entropy scores. We then select reasoning tokens by
\begin{equation}
    w_i^{\mathrm{rea}}
    =
    \mathbf{1}\!\left[
        i\in\operatorname{TopK}_{\rho_r}
        (\bar H_{\mathcal{N}_{\mathrm{vis}}})
    \right].
\end{equation}
This strict routing keeps the visual and reasoning branches disjoint.
Unselected tokens remain anchored to the original trajectory and receive no
direct policy reward.

\subsection{Inference Procedure}

Algorithm~\ref{alg:td-ltts} summarizes our inference procedure. Starting from
the model's own chain-of-thought rollout, we refine an early latent prefix and
use decoded candidates as reward feedback for subsequent updates. Candidate
sets combine one greedy continuation with stochastic continuations from lightly
perturbed latent prefixes, diversifying search while keeping the VLM frozen.

\begin{algorithm}[t]
\small
\caption{Token-Disentangled Latent Test-Time Scaling}
\label{alg:td-ltts}
\begin{algorithmic}[1]
\REQUIRE Multimodal context $\xi=(I,q)$, frozen VLM $p_\theta$, update steps $K$
\STATE Generate initial rollout $y$ and hidden states $h_{1:T}$
\STATE Initialize editable prefix $z^{(0)}=[h_1,\ldots,h_L]$
\STATE Compute image-sensitivity scores $\{v_i\}_{i=1}^{L}$
\STATE Set $\mathcal{X}^{(0)}\!=\!\{y\}$, $x_0^\star\!=\!y$; compute $R_{\mathrm{rea}}(y)$ (with $R_{\mathrm{vis}}(y)\!=\!0$ by construction)
\FOR{$k=0,\ldots,K-1$}
    \STATE Compute entropy scores $\{H_i\}_{i=1}^{L}$ from $z^{(k)}$
    \STATE Select $w^{\mathrm{vis}}$ by image-sensitivity $\operatorname{TopK}_{\rho_v}$
    \STATE Select $w^{\mathrm{rea}}$ by standardized entropy $\operatorname{TopK}_{\rho_r}$ among non-visual tokens
    \STATE Update $z^{(k+1)}$ using $R_{\mathrm{rea}}(x_k^{\star})$ and $R_{\mathrm{vis}}(x_k^{\star})$
    \STATE Decode $\mathcal{X}^{(k+1)}$ from $z^{(k+1)}$ and noisy variants of $z^{(k+1)}$
    \STATE Select $x_{k+1}^\star=\arg\max_{x\in\mathcal{X}^{(k+1)}}R_{\mathrm{rea}}(x)$
    \STATE Compute $R_{\mathrm{rea}}(x_{k+1}^\star)$ and $R_{\mathrm{vis}}(x_{k+1}^\star)$ for the next update
\ENDFOR
\RETURN final answer aggregated from $\{y\}\cup\mathcal{X}^{(1)}\cup\cdots\cup\mathcal{X}^{(K)}$
\end{algorithmic}
\end{algorithm}

\section{Experiment}

\begin{table*}[t]
\caption{Main results on multimodal reasoning benchmarks. Best results in each model block are highlighted in \textbf{bold}, and second-best results are underlined.}
\vspace{1mm}
\centering
\small
\setlength{\tabcolsep}{5pt}
\resizebox{\textwidth}{!}{
\begin{tabular}{lcccc|cccc}
\toprule
\multirow{2}{*}{\textbf{Method}}
& \multicolumn{4}{c|}{\textbf{Perception}}
& \multicolumn{4}{c}{\textbf{Reasoning}} \\
\cmidrule(lr){2-5} \cmidrule(lr){6-9}
& \textbf{MMStar} & \textbf{RWQA} & \textbf{Hallusion} & \textbf{Avg.}
& \textbf{ScienceQA} & \textbf{MathVista} & \textbf{LogicVista} & \textbf{Avg.} \\
\midrule
\multicolumn{9}{c}{\textit{\textbf{Qwen2.5-VL-7B}}} \\
\midrule
CoT              & 62.00 & 63.66 & 70.56 & 65.41 & 89.74 & 67.80 & \underline{44.52} & 67.35 \\
Self-consistency & 63.47 & 65.49 & 68.77 & 65.91 & 89.60 & 70.90 & 42.51 & 67.67 \\
Best-of-N        & 63.93 & 66.37 & 70.56 & 66.95 & \underline{90.63} & \underline{71.60} & 42.73 & 68.32 \\
Reward-only      & \underline{64.67} & \underline{67.06} & 70.45 & \underline{67.39} & 90.12    & 70.90 & 42.95 & 67.99 \\
\midrule
LatentSeek(reasoning)       & 61.60 & 63.79 & 69.09 & 64.83 & 88.29 & 66.50 & 44.07 & 66.29 \\
LatentSeek(perception) & 62.47 & 62.35 & 68.24 & 64.35 & 90.53 & 70.20 & 44.30 & \underline{68.34} \\
DMLR                   & 60.10 & 64.17 & \underline{70.80} & 65.02 & 89.14 & 69.10 & 43.15 & 67.13 \\
\rowcolor{orange!15}\textbf{Ours}
                 & \textbf{64.73} & \textbf{67.45} & \textbf{71.71} & \textbf{67.96}
                 & \textbf{90.73} & \textbf{72.10} & \textbf{46.98} & \textbf{69.94} \\

\midrule
\multicolumn{9}{c}{\textit{\textbf{InternVL3.5-8B}}} \\
\midrule
CoT              & 67.13 & 64.36 & 66.04 & 65.84 & 95.04 & 69.80 & 45.19 & 70.01 \\
Self-consistency & 67.13 & 64.71 & 66.25 & 66.03 & 95.64 & 70.10 & 45.90 & 70.55 \\
Best-of-N        & 67.80 & 64.71 & \underline{66.56} & 66.36 & 94.75 & \underline{72.10} & \textbf{46.98} & 71.28 \\
Reward-only      & \underline{68.07} & \underline{65.10} & \textbf{66.61} & \underline{66.59} & \underline{96.98} & 71.30 & 46.53 & \underline{71.60} \\
\midrule
LatentSeek(reasoning)       & 67.07 & 61.31 & 63.51 & 63.96 & 94.65 & 70.60 & 45.64 & 70.30 \\
LatentSeek(perception)     & 66.20 & 62.61 & 64.04 & 64.28 & 91.08 & 72.00 & 44.97 & 69.35 \\
DMLR                       & 67.47 & 63.19 & 65.65 & 65.44 & 95.44 & 71.80 & 45.96 & 71.07 \\
\rowcolor{blue!15}\textbf{Ours}
                 & \textbf{68.80} & \textbf{65.23} & 66.46 & \textbf{66.83}
                 & \textbf{97.08} & \textbf{72.30} & \underline{46.76} & \textbf{72.05} \\

\bottomrule
\end{tabular}
}
\vspace{1mm}
\label{tab:main_results}
\end{table*}

\subsection{Experiment Setting}

\paragraph{Backbones.}
We evaluate our method on two main open-source MLLM backbones,
Qwen2.5-VL-7B-Instruct~\cite{bai2025qwen2} and
InternVL3.5-8B-Instruct~\cite{wang2025internvl3}, and further test
generality across six additional backbones from different families and
post-training recipes in Table~\ref{tab:detailed_analysis}. In all
settings the backbone parameters are frozen, and our method only
optimizes instance-specific latent states at test time.

\paragraph{Benchmarks.}
We evaluate on benchmarks covering two capability groups. For perception
and visually grounded understanding, we use
MMStar~\cite{mmstarchen2024we}, RealWorldQA, and
HallusionBench~\cite{guan2024hallusionbench}. For reasoning-centric
evaluation, we use ScienceQA-IMG~\cite{scienceqalu2022learn},
MathVista~\cite{lu2023mathvista}, and
LogicVista~\cite{xiao2024logicvista}. All methods use the same answer
extraction and evaluation protocol.

\paragraph{Baselines.}
We compare against test-time baselines using the same frozen backbone.
\textbf{CoT}~\cite{wei2022chain} uses chain-of-thought prompting.
\textbf{Self-consistency}~\cite{wang2022self} samples multiple responses
and aggregates final answers, while \textbf{Best-of-N}~\cite{snell2024scaling}
selects the highest-scoring candidate. \textbf{Reward-only} applies our
reasoning reward only for output-space reranking, without latent updates
or token routing. \textbf{LatentSeek}~\cite{li2025seek} is a prior
latent-adaptation baseline; we use its reasoning- and perception-reward
variants, denoted \textbf{LatentSeek(reasoning)} and
\textbf{LatentSeek(perception)}. \textbf{DMLR}~\cite{liu2025reasoning}
is a multimodal latent-refinement method that emphasizes
preserving visual information during latent computation. Together,
these baselines separate our method from output-space scaling and prior
latent-space refinement.

\paragraph{Implementation Details.}
Unless otherwise noted, we use $K=4$ latent refinement steps as the
default. To equalize decoding cost, all sampling-based baselines are
allocated $N=16$ candidate continuations, matching the total number
of decoded candidates in our method. Our method additionally incurs
reward-scoring and image-sensitivity forward passes; full hyperparameters,
decoding settings, and per-call budget accounting are deferred to
Appendices~\ref{sec:appendix-implementation},
\ref{sec:appendix-default-config},
and~\ref{sec:appendix-ablation-matched-compute}.

\subsection{Main Results}

Table~\ref{tab:main_results} reports the main comparison on
Qwen2.5-VL-7B and InternVL3.5-8B. Our method achieves the best average
on both perception and reasoning groups across both backbones, lifting
the Qwen2.5-VL-7B averages by $+2.55$/$+2.59$ and the InternVL3.5-8B
counterparts by $+0.99$/$+2.04$ over CoT. Because the backbone is
frozen, these gains are obtained purely through test-time computation
on the latent states and are consistent across both architectures and
capability groups, suggesting that the improvement comes from
instance-level trajectory refinement rather than any backbone-specific
bias.

Against output-space baselines under matched decoded-candidate budgets,
our method yields stronger group averages on both backbones: on
Qwen2.5-VL-7B the reasoning margins over Best-of-N and Reward-only are
$+1.62$ and $+1.95$, and on InternVL3.5-8B they are $+0.77$ and
$+0.45$, with consistent positive gains on perception. Editing the
latent that drives subsequent decoding uses the candidate budget more
efficiently than sampling-and-reranking, as the refined latent commits
the improvement to every later token rather than only the selected
candidate. The LatentSeek comparison is sharper: on Qwen2.5-VL-7B both
variants fall \emph{below} CoT on perception ($64.83$/$64.35$ vs.\
$65.41$), and the reasoning-reward variant additionally drops reasoning
($66.29$ vs.\ $67.35$), showing that a uniform rule-based reward can
degrade a competent rollout because a reasoning-only signal erodes
perception-sensitive tokens and vice versa. Token-routed refinement
avoids this regression by dispatching different reward signals to
different token roles, retaining the strongest performance on both
groups simultaneously.

\Needspace{22\baselineskip}
\subsection{Detailed Analysis}

\begin{table*}[t]
\caption{Performance comparison on perception and reasoning benchmarks.
\textcolor{mygreen}{$\uparrow$} and \textcolor{myred}{$\downarrow$} indicate
performance changes compared with the base model.}
\label{tab:detailed_analysis}
\centering
\small
\setlength{\tabcolsep}{2pt}
\renewcommand{\arraystretch}{1.15}
\resizebox{\textwidth}{!}{
\begin{tabular}{l|cccc|cccc}
\toprule
\multirow{2}{*}{\textbf{Base Model}}
& \multicolumn{4}{c|}{\textbf{Perception}}
& \multicolumn{4}{c}{\textbf{Reasoning}} \\
\cmidrule(lr){2-5} \cmidrule(lr){6-9}
& \textbf{MMStar} & \textbf{RWQA} & \textbf{Hallusion} & \textbf{Avg.}
& \textbf{ScienceQA} & \textbf{MathVista} & \textbf{LogicVista} & \textbf{Avg.} \\
Qwen2.5-VL-3B~\citep{bai2025qwen2}
& 59.50 & 54.07 & 63.30 & 58.96
& 80.76 & 63.14 & 40.93 & 61.61 \\
\rowcolor{mygray}
\textbf{Ours}
& 62.20 \inc{2.70} & 56.13 \inc{2.06} & 65.30 \inc{2.00} & 61.21 \inc{2.25}
& 80.91 \inc{0.15} & 64.05 \inc{0.91} & 41.88 \inc{0.95} & 62.28 \inc{0.67} \\
\midrule
InternVL3.5-4B~\citep{wang2025internvl3}
& 69.10 & 64.87 & 63.62 & 65.86
& 93.75 & 54.77 & 41.36 & 63.29 \\
\rowcolor{mygray}
\textbf{Ours}
& 70.70 \inc{1.60} & 65.13 \inc{0.26} & 64.04 \inc{0.42}& 66.62 \inc{0.76}
& 94.86 \inc{1.11} & 63.53 \inc{8.76} & 43.04 \inc{1.68} & 67.14 \inc{3.85} \\
\midrule
LLaVA-OV-1.5-8B~\citep{an2025llava}
& 62.67 & 66.80 & 61.72 & 63.73
& 90.68 & 68.30 & 43.85 & 67.61 \\
\rowcolor{mygray}
\textbf{Ours}
& 65.80 \inc{3.13} & 66.67 \dec{0.13} & 64.88 \inc{3.15} & 65.78 \inc{2.05}
& 96.43 \inc{5.75} & 70.40 \inc{2.10} & 46.53 \inc{2.68} & 71.12 \inc{3.51} \\
\midrule
MiMO-VL-RL-8B~\citep{li2025xiaomi}
& 48.40 & 45.36 & 49.11 & 47.62
& 86.71 & 63.10 & 37.14 & 62.32 \\
\rowcolor{mygray}
\textbf{Ours}
& 51.60 \inc{3.20} & 57.39 \inc{12.03} & 58.36 \inc{9.25} & 55.78 \inc{8.16}
& 93.85 \inc{7.14} & 66.20 \inc{3.10} & 48.55 \inc{11.41} & 69.53 \inc{7.22} \\
\midrule
Qwen3-VL-8B
& 71.33 & 69.28 & 67.19 & 69.27
& 93.11 & 77.10 & 48.55 & 72.92 \\
\rowcolor{mygray}
\textbf{Ours}
& 71.73 \inc{0.40} & 68.89 \dec{0.39} & 68.45 \inc{1.26} & 69.69 \inc{0.42}
& 94.89 \inc{1.78} & 76.90 \dec{0.20} & 52.35 \inc{3.80} & 74.71 \inc{1.79} \\
\midrule
Qwen2.5-VL-32B~\citep{bai2025qwen2}
& 67.9 & 67.8 & 65.3 & 67.0
& 93.2 & 77.4 & 50.3 & 73.6 \\
\rowcolor{mygray}
\textbf{Ours}
& 68.0 \inc{0.1} & 69.7 \inc{1.9} & 69.7 \inc{4.4} & 69.1 \inc{2.1}
& 93.6 \inc{0.4} & 77.8 \inc{0.4} & 51.7 \inc{1.4} & 74.4 \inc{0.7} \\
\bottomrule
\end{tabular}
}
\end{table*}

\leavevmode
\begin{wrapfigure}{r}{0.48\textwidth}
\centering
\includegraphics[width=\linewidth]{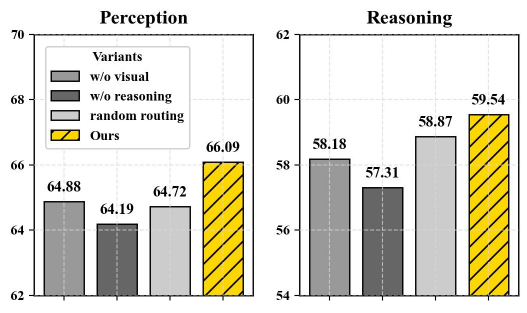}
\caption{Component ablation. Perception is the average of MMStar and
RealWorldQA; reasoning is the average of MathVista and LogicVista
(two representative tasks per group, used as a cost-controlled subset
of the Table~\ref{tab:main_results} groups).}
\label{fig:component_ablation}
\end{wrapfigure}
\textbf{[Ana.1] Component Ablation.}
We ablate the two reward branches and the token-routing module with
other settings fixed; Figure~\ref{fig:component_ablation} reports
\emph{two-task representative averages} (perception over MMStar+RWQA,
reasoning over MathVista+LogicVista) as a cost-controlled subset of
the Table~\ref{tab:main_results} groups.
Removing the visual reward ($\text{w/o}~R_{\mathrm{vis}}$) drops both
averages from $66.09$/$59.54$ to $64.88$/$58.18$, while removing the
reasoning reward ($\text{w/o}~R_{\mathrm{rea}}$) causes the largest
drop ($64.19$/$57.31$), identifying bounded-likelihood scoring as the
dominant signal and image-token engagement as the necessary corrector
for perception-sensitive tokens. Replacing token-aware routing with
random routing also degrades both groups ($64.72$/$58.87$), confirming
that multiple rewards alone are insufficient without role-aware token
assignment. Extended sweeps over reward formulations and routing
signals appear in
Appendices~\ref{sec:appendix-ablation-reward}
and~\ref{sec:appendix-ablation-routing-signal}; the routing budget $\rho$
sweep and strict-vs-overlap mask comparison show that hard top-$K$
selection is robust within the explored range.

\noindent \textbf{[Ana.2] Generality Across Backbones.}
Table~\ref{tab:detailed_analysis} extends the evaluation to six
additional backbones covering compact (Qwen2.5-VL-3B,
InternVL3.5-4B), 8B (LLaVA-OneVision-1.5-8B, MiMO-VL-RL-8B,
Qwen3-VL-8B), and large (Qwen2.5-VL-32B) models from different
families and post-training recipes. Our method improves both group
averages on all six backbones. The largest
improvements appear on the RL-post-trained MiMO-VL-RL-8B, while stronger backbones such as
Qwen3-VL-8B show smaller gains as the initial rollout already
approaches benchmark headroom; group averages remain positive even
when isolated cells drop within the noise band (at most $-0.39$).
The consistency across model family, scale (3B--32B), and post-training
recipe (SFT vs.\ RL) suggests that the gains arise from
instance-level latent refinement rather than any architectural
assumption.

\Needspace{24\baselineskip}
\begin{wrapfigure}[16]{r}{0.48\textwidth}
\centering
\includegraphics[width=\linewidth]{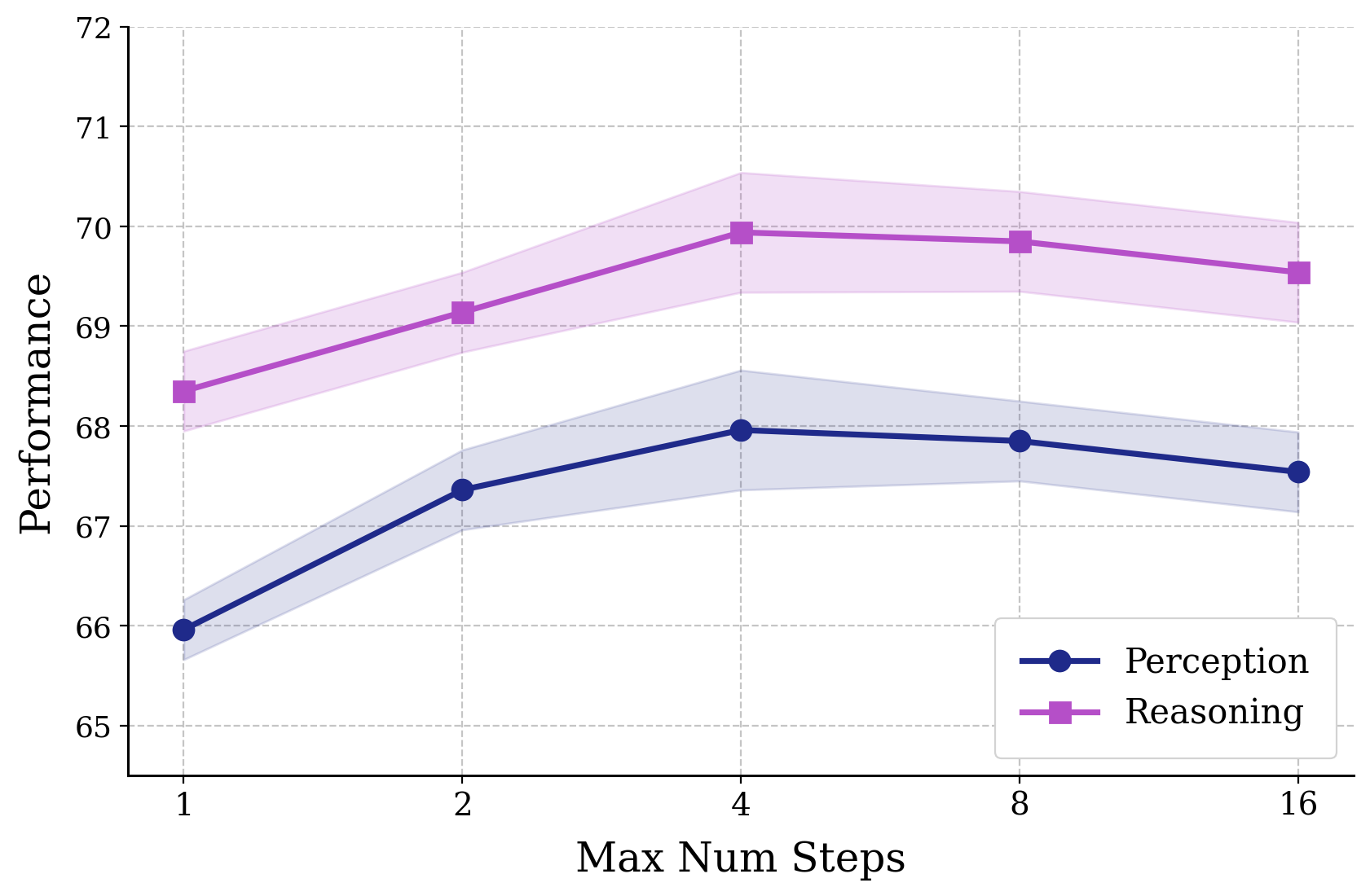}
\caption{Effect of the maximum number of latent refinement steps $K$ on the
three-benchmark perception and reasoning averages from
Table~\ref{tab:main_results}.}
\label{fig:step_scaling}
\end{wrapfigure}

\noindent \textbf{[Ana.3] Test-Time Scaling and Token Budget.} We study how performance scales with the maximum number of latent
refinement steps $K$. Figure~\ref{fig:step_scaling} reports the
perception/reasoning averages for $K \in \{1, 2, 4, 8, 16\}$. Both
groups improve monotonically from $K=1$ to $K=4$ (perception
$65.96\!\to\!67.36\!\to\!67.96$; reasoning
$68.34\!\to\!69.13\!\to\!69.94$), and then \emph{saturate} at $K\!\geq\!8$
within the noise band (perception $67.86$/$67.74$; reasoning
$69.88$/$69.78$, with overlapping confidence intervals across
$K\!\in\!\{4,8,16\}$); the same $K=4$ optimum appears in the per-benchmark
sweep in Appendix~\ref{sec:appendix-ablation-optim}. This indicates
that a small number of latent updates suffices to correct the initial
trajectory and that further refinement yields no additional measurable
gain, so we adopt $K=4$ as the compute-optimal default in the main
results.

\section{Conclusion}

We introduced \textbf{Token-Disentangled Latent Test-Time Scaling}, a
role-aware latent refinement framework for frozen multimodal large language
models. By routing visual feedback to image-sensitive tokens and reasoning
feedback to uncertain decision tokens, our method avoids applying a single global
reward to all editable latent states. Experiments show consistent improvements
over CoT and over test-time scaling baselines, while
ablations confirm the importance of both reward decomposition and token-aware
routing. Future work can further reduce reward-estimation cost and extend this
framework to longer multimodal interactions.

\section*{Limitations}

Our method improves frozen MLLMs at test time, but it also introduces additional
inference cost because each example requires latent refinement and multiple
candidate continuations. The reward signals are proxy objectives rather than
guaranteed correctness verifiers, so visual engagement and likelihood-based
reasoning scores can be noisy for ambiguous images, underspecified questions, or
answers requiring external knowledge; the bounded $\tanh$ in $R_{\mathrm{vis}}$
limits but does not eliminate this. Token routing uses a hard top-$K$ selection
for branch disjointness; soft or learned routing is left to future work. In addition, our method requires open-weight backbones whose hidden
states can be edited and differentiated through, which excludes
API-only closed models by construction. Extending the framework to
longer multi-turn settings and broader real-world domains remains
future work.

\section*{Acknowledgments}

This work was supported by the National Key R\&D Program of China under Grant
No.~2024YFE0202800 and the National Natural Science Foundation of China under
Grant Nos.~62522605 and 62376118. Long Chen was additionally supported by the
National Natural Science Foundation of China under Grant Nos.~62522216 and
62402408, and the Hong Kong SAR Research Grants Council under Grant
Nos.~26208924 and 16219025.

\FloatBarrier
\bibliographystyle{conference}
\bibliography{custom}

\clearpage
\appendix
\section*{Appendix Overview}
This appendix is organized as follows:
\begin{itemize}
    \item[\textbf{A.}] We define the initial-rollout visual baseline used by
    the visual reward.
    \item[\textbf{B.}] We describe implementation details, including the
    reasoning reward, default configuration, latent editing, and candidate
    decoding.
    \item[\textbf{C.}] We report supplementary ablation studies covering
    routing budgets, routing signals, reward components, latent
    optimization hyperparameters, and matched-budget comparisons.
    \item[\textbf{D.}] We provide qualitative case studies illustrating how
    visual and reasoning rewards are routed to different latent-token
    groups, along with a representative failure case.
\end{itemize}

\section{Initial-Rollout Visual Baseline}
\label{sec:appendix-visual-baseline}

The visual reward compares the image-token attention of a decoded candidate
against the initial rollout from the same example. Let $y$ be the initial
chain-of-thought trajectory and $E(\cdot)$ the trajectory-level visual
engagement score defined in Section~\ref{sec:method}. For a later candidate
$x$, the reward is
\begin{equation}
    R_{\mathrm{vis}}(x)
    =
    \tanh\!\left(
        \frac{E(x)-E(y)}{\tau_{\mathrm{vis}}}
    \right).
\end{equation}
The initial rollout is therefore the per-instance baseline, so
$R_{\mathrm{vis}}(y)=0$. Concretely, we set $\mathcal{X}^{(0)}=\{y\}$ so the
first latent update is driven by the reasoning reward of $y$ alone, without
any extra decoding at $z^{(0)}$. From the first update onward, each decoded
candidate set $\mathcal{X}^{(k)}$ ($k\ge 1$) comes from a perturbed latent
prefix, so its candidates have engagement scores that differ from $E(y)$,
and the visual branch receives nonzero relative feedback whenever the
selected candidate attends to image tokens more or less strongly than the
initial rollout. This signal is used only for image-sensitive latent tokens
and is not treated as an answer-correctness verifier.

\section{Implementation Details}
\label{sec:appendix-implementation}

\subsection{Reasoning Reward Details}
\label{sec:appendix-reasoning-reward}

For a normalized short prediction $c$, the text-prior-corrected score is
\begin{equation}
    S(c)
    =
    \log p_\theta(c\mid I,q)
    -
    \lambda_{\mathrm{p}}
    \log p_\theta(c\mid q).
\end{equation}
We map this score to a bounded scalar and center it with a per-step
baseline:
\begin{equation}
    r(c)=\tanh\!\left(S(c)/\tau_s\right),
    \qquad
    \bar r(x)=r(c(x))-b_s.
\end{equation}
For the relative margin term, let $\mathcal{B}(c)$ denote the competing
predictions from the same refinement step, including the initial rollout
when available. We compute
\begin{equation}
    \begin{aligned}
    \Delta_{\mathrm{mar}}(c)
    &=
    S(c)
    -
    \max_{c'\in\mathcal{B}(c)} S(c'), \\
    r_{\mathrm{mar}}(c)
    &=
    \tanh\!\left(\Delta_{\mathrm{mar}}(c)/\tau_{\mathrm{mar}}\right).
    \end{aligned}
\end{equation}

\subsection{Default Configuration}
\label{sec:appendix-default-config}

Unless otherwise specified, all reported results use the configuration
below.

\begin{itemize}\itemsep1pt
    \item \textbf{Refinement schedule.} $K=4$ refinement steps; editable
    prefix length $L=\min(\lfloor\rho T\rfloor,300)$ with start position
    $s=0$ and $\rho=0.5$; Adam optimizer with learning rate $0.05$.
    \item \textbf{Token routing.} Visual budget $\rho_v=0.4$ on editable
    tokens by image sensitivity; reasoning budget $\rho_r=0.4$ on
    non-visual tokens by standardized LM-head entropy; strictly disjoint
    masks.
    \item \textbf{Visual reward.} $\tau_{\mathrm{vis}}=0.2$, final attention
    layer, engagement scored over at most $128$ generated tokens. Image
    sensitivity uses raw-image patch blackening with patch size $14$ and
    drop probability $0.5$.
    \item \textbf{Reasoning reward.} Bounded-score weight $\lambda_s=1.0$,
    margin weight $\lambda_{\mathrm{mar}}=0.5$, text-prior weight
    $\lambda_{\mathrm{p}}=1.0$; visual and reasoning policy weights
    $\lambda_{\mathrm{pg}}^{\mathrm{vis}}=\lambda_{\mathrm{pg}}^{\mathrm{rea}}=1.0$.
    \item \textbf{Regularization.} Anchor weight $\lambda_{\mathrm{anchor}}=0.05$;
    logit-entropy weight $\lambda_{\mathrm{ent}}=0.01$.
    \item \textbf{LatentSeek baseline.} A separate configuration closer to
    its original setting: $10$ latent update steps, $\rho=0.2$, learning
    rate $0.05$.
\end{itemize}

\subsection{Latent Editing and Candidate Decoding}
\label{sec:appendix-latent-editing}

The editable states are the last decoder-layer hidden states of the
generated response, immediately before the frozen LM head. They are
collected from the model's native generation API with
\texttt{return\_\allowbreak dict\_\allowbreak in\_\allowbreak generate=True} and
\texttt{output\_\allowbreak hidden\_\allowbreak states=True}. We do not edit visual-encoder states,
model weights, or intermediate decoder layers.

The optimized continuous states are not directly inserted into the
transformer's KV cache. At each refinement step, each edited state is first
projected through the frozen LM head to obtain a latent-induced prefix
distribution. We decode one greedy prefix by taking the argmax token at
each editable position, and decode three noisy-prefix variants by adding
Gaussian noise with base standard deviation $0.3$ to the edited states
before top-$k$ sampling from the LM-head logits. Each resulting discrete
prefix is concatenated with the original multimodal prompt and unedited
response prefix before the model continues autoregressive generation with
its native \texttt{generate()} function.

Every candidate continuation is generated from a fresh generation call. We
reset the model's multimodal generation state before each call and do not
reuse the KV cache from the initial rollout or from previous candidates.
Multimodal generation inputs such as image tensors, image-grid metadata,
image flags, and model-specific token-type ids are copied from the original
prompt inputs and extended when needed to match the latent-induced prefix
length.

Across the default $K=4$ refinement steps the method decodes $4$ candidates
per step (one greedy and three noisy); the original rollout is also kept as
a reference candidate for scoring and final selection. Short answers are
normalized by parsing structured outputs when available (JSON fields such
as \texttt{final answer}, \texttt{final\_answer}, or \texttt{answer}) and by
applying the benchmark adapter's answer parser for multiple-choice and
pairwise-evaluation tasks. The default final aggregation groups candidates
by normalized short answer, selects the group with the largest count, and
breaks ties by the highest reward within the group; equivalently, the
implementation uses the composite score $10\cdot\mathrm{count}+\max R$ per
normalized-answer group. A reward-weighted log-sum-exp aggregation is also
implemented but is not the default.

For model-specific hidden-state handling, Qwen-style models use the
HuggingFace generation output directly. InternVL models are wrapped with a
lightweight generation adapter so that their \texttt{generate()} output
exposes the same prompt-plus-generation sequence convention expected by the
latent-prefix code. This keeps hidden-state collection, prefix construction,
and candidate decoding identical across the supported backbones.

\section{Additional Ablation Studies}
\label{sec:appendix-additional-ablations}

All ablations use Qwen2.5-VL-7B as the frozen backbone and the default
configuration unless otherwise specified. Results are reported as accuracy
(\%). We use LogicVista and RealWorldQA as representative reasoning-heavy
and perception-heavy benchmarks; the matched-budget comparison
(Section~\ref{sec:appendix-ablation-matched-compute}) additionally covers
MathVista. The baseline rollout values are $44.52$ on LogicVista and
$63.66$ on RealWorldQA, matching the CoT entries in
Table~\ref{tab:main_results}. Where shown, parenthetical values report the
absolute improvement in percentage points over this baseline, computed
from unrounded scores prior to rounding, and Macro denotes the
two-benchmark average.

\subsection{Token Routing Budget}
\label{sec:appendix-ablation-routing-budget}

We sweep the visual and reasoning routing budgets $\rho_v$ and $\rho_r$ and
test disjoint vs.\ overlapping masks and swapped reward-token assignment.

\begin{table}[!htbp]
\centering
\small
\caption{Diagonal top-$k$ ratio sweep with $\rho_v=\rho_r$.}
\label{tab:ablation-routing-budget}
\begin{tabular}{lcc}
\toprule
$\rho_v=\rho_r$ & LogicVista & RealWorldQA \\
\midrule
$0.10$ & 45.10 (+0.58) & 66.40 (+2.74) \\
$0.20$ & 45.55 (+1.03) & 66.80 (+3.14) \\
$0.30$ & 46.31 (+1.79) & 67.19 (+3.53) \\
$0.40$ & \textbf{46.98 (+2.46)} & \textbf{67.45 (+3.79)} \\
$0.50$ & 46.53 (+2.01) & 67.06 (+3.40) \\
$0.60$ & 45.86 (+1.34) & 66.80 (+3.14) \\
$0.80$ & 45.41 (+0.89) & 66.40 (+2.74) \\
\bottomrule
\end{tabular}
\end{table}

The routing budget is not monotonic: both benchmarks peak at $\rho=0.4$.
Larger budgets such as $0.6$ and $0.8$ as well as very small budgets such
as $0.1$ both reduce the gains, but all sweep settings remain above the
baseline, indicating that the method is robust to threshold choice within
the explored range.

\begin{table}[!htbp]
\centering
\small
\caption{Strict disjoint masks vs.\ overlapping masks at $\rho=0.4$.}
\label{tab:ablation-routing-strict-vs-overlap}
\begin{tabular}{lcc}
\toprule
Setting & LogicVista & RealWorldQA \\
\midrule
strict disjoint (default) & \textbf{46.98 (+2.46)} & \textbf{67.45 (+3.79)} \\
overlap allowed & 46.31 (+1.79) & 67.06 (+3.40) \\
\bottomrule
\end{tabular}
\end{table}

Strict disjoint routing is stronger than overlap on both benchmarks,
supporting our default of enforcing disjoint visual and reasoning masks.

\begin{table}[!htbp]
\centering
\small
\caption{Decoupled visual and reasoning routing budgets.}
\label{tab:ablation-routing-asymmetric}
\adjustbox{max width=\linewidth}{
\begin{tabular}{llcc}
\toprule
$\rho_v$ & $\rho_r$ & LogicVista & RealWorldQA \\
\midrule
$0.20$ & $0.20$ & 45.10 (+0.58) & 66.27 (+2.61) \\
$0.20$ & $0.40$ & 45.41 (+0.89) & 66.40 (+2.74) \\
$0.20$ & $0.60$ & 45.19 (+0.67) & 66.53 (+2.87) \\
$0.20$ & $0.80$ & 44.97 (+0.45) & 66.27 (+2.61) \\
$0.40$ & $0.20$ & 46.31 (+1.79) & 66.93 (+3.27) \\
$0.40$ & $0.40$ & \textbf{46.98 (+2.46)} & \textbf{67.45 (+3.79)} \\
$0.40$ & $0.60$ & 46.76 (+2.24) & 67.19 (+3.53) \\
$0.40$ & $0.80$ & 46.31 (+1.79) & 67.06 (+3.40) \\
$0.60$ & $0.20$ & 45.86 (+1.34) & 66.80 (+3.14) \\
$0.60$ & $0.40$ & 45.86 (+1.34) & 66.67 (+3.01) \\
$0.60$ & $0.60$ & 45.41 (+0.89) & 66.40 (+2.74) \\
$0.60$ & $0.80$ & 45.19 (+0.67) & 66.40 (+2.74) \\
$0.80$ & $0.20$ & 45.10 (+0.58) & 66.27 (+2.61) \\
$0.80$ & $0.40$ & 45.10 (+0.58) & 66.27 (+2.61) \\
$0.80$ & $0.60$ & 45.19 (+0.67) & 66.40 (+2.74) \\
$0.80$ & $0.80$ & 45.19 (+0.67) & 66.40 (+2.74) \\
\bottomrule
\end{tabular}
}
\end{table}

The decoupled sweep shows that the symmetric setting
$\rho_v=\rho_r=0.4$ is best on both benchmarks. Nearby asymmetric settings
such as $\rho_v=0.4,\rho_r=0.6$ remain competitive, indicating that the
optimum is reasonably flat around the default.

\begin{table}[!htbp]
\centering
\small
\caption{Swapped-routing sanity check at $\rho=0.4$.}
\label{tab:ablation-routing-swapped}
\begin{tabular}{lcc}
\toprule
Setting & LogicVista & RealWorldQA \\
\midrule
correctly routed (default) & \textbf{46.98 (+2.46)} & \textbf{67.45 (+3.79)} \\
swapped routing & 45.41 (+0.89) & 66.27 (+2.61) \\
\bottomrule
\end{tabular}
\end{table}

Swapped routing is clearly weaker than the correctly routed strict setting
on both benchmarks, supporting the claim that the direction of
reward-token assignment matters rather than gains coming from updating an
arbitrary subset of tokens.

\subsection{Routing Signal}
\label{sec:appendix-ablation-routing-signal}

We compare the default image-sensitivity routing signal against
alternative visual routing signals, and vary the reasoning-side signal
while holding the visual channel fixed. We additionally vary the patch
size and drop probability used for raw-image patch blackening.

\begin{table}[!htbp]
\centering
\small
\caption{Visual routing signal ablation.}
\label{tab:ablation-routing-signal}
\adjustbox{max width=\linewidth}{
\begin{tabular}{lcccc}
\toprule
Setting & LogicVista & RealWorldQA & Macro & $\Delta$ \\
\midrule
image-sensitivity top-$k$ (default) & \textbf{46.98} & \textbf{67.45} & \textbf{57.22} & \textbf{+3.13} \\
visual-focus only & 46.31 & 67.19 & 56.75 & +2.66 \\
hard top-$k$ visual focus & 46.53 & 67.06 & 56.80 & +2.71 \\
cosine-prototype score & 45.41 & 66.93 & 56.17 & +2.08 \\
\bottomrule
\end{tabular}
}
\end{table}

Image-sensitivity top-$k$ outperforms prototype-similarity and
visual-focus alternatives, supporting it as the default visual routing
signal.

\begin{table}[!htbp]
\centering
\small
\caption{Reasoning routing signal ablation.}
\label{tab:ablation-routing-signal-reason}
\adjustbox{max width=\linewidth}{
\begin{tabular}{lcccc}
\toprule
Setting & LogicVista & RealWorldQA & Macro & $\Delta$ \\
\midrule
neg.\ log-prob (default) & \textbf{46.98} & \textbf{67.45} & \textbf{57.22} & \textbf{+3.13} \\
token loss & 46.76 & 67.19 & 56.97 & +2.88 \\
random non-visual & 45.41 & 66.40 & 55.90 & +1.81 \\
\bottomrule
\end{tabular}
}
\end{table}

Both likelihood-based routing signals clearly outperform random
non-visual routing, indicating that reasoning tokens cannot be selected
arbitrarily. Negative log-probability is slightly stronger than token
loss and is used as the default reasoning-side routing signal.

\begin{table}[!htbp]
\centering
\small
\caption{Patch-blackening robustness for image-sensitivity routing.
Values are two-benchmark Macro averages.}
\label{tab:ablation-corruption}
\adjustbox{max width=\linewidth}{
\begin{tabular}{lccc}
\toprule
Patch / Drop & 0.25 & 0.50 & 0.75 \\
\midrule
patch=$7$  & 56.50 & 56.75 & 56.95 \\
patch=$14$ & 56.40 & \textbf{57.22} & 56.80 \\
patch=$28$ & 56.60 & 56.50 & 56.70 \\
patch=$56$ & 56.60 & 56.85 & 56.20 \\
\bottomrule
\end{tabular}
}
\end{table}

All patch/drop settings remain above Macro $56.20$ (vs.\ base Macro
$54.09$), so image-sensitivity routing is not overly fragile. The default
(patch size $14$, drop $0.50$) reaches Macro $57.22$. Large patches with
high drop probability are the weakest combination, suggesting that overly
strong image corruption makes the sensitivity signal noisy.

\begin{table*}[!tbp]
\centering
\small
\caption{Readable span-level visualization of token routing for the
MMStar dog-counting case. Color legend follows the surrounding paragraph.}
\label{tab:case-study-token-routing}
\begin{tabular}{p{0.16\textwidth}p{0.76\textwidth}}
\toprule
Field & Routed rollout excerpt \\
\midrule
Question &
How many dogs can be seen in the image? Options: A: $3$, B: $2$,
C: $1$, D: $4$. \\
\midrule
Initial rollout &
\anchortoken{I need to identify all the dogs in the image.}
\anchortoken{There is}
\vistok{one dog visible}
\vistok{on the left side}
\anchortoken{of the image, lying down.}
\reatok{No other dogs are clearly visible}
\anchortoken{in the rest of the room.}
Final answer: C. \\
\midrule
Selected refined candidate &
\anchortoken{I need to carefully identify the dogs in the image.}
\vistok{There is a dog visible on the left side}
\anchortoken{of the image, lying down.}
\vistok{Another dog is partially visible behind the couch}
\vistok{near the center of the room.}
\reatok{No other dogs are clearly visible.}
Final answer: B. \\
\midrule
Routing signal &
Image-sensitivity top-$k$/mean $=1.79/0.32$; entropy top-$k$/mean
$=1.83/0.54$. \\
\bottomrule
\end{tabular}
\end{table*}

\subsection{Reward Components}
\label{sec:appendix-ablation-reward}

We ablate the visual reward formulation, the reasoning-reward components,
the visual-reward temperature $\tau_{\mathrm{vis}}$, and the attention
layer used for engagement scoring.

\begin{table}[!htbp]
\centering
\small
\caption{Visual reward ablation.}
\label{tab:ablation-visual-reward}
\adjustbox{max width=\linewidth}{
\begin{tabular}{lccc}
\toprule
Setting & LogicVista & RealWorldQA & Macro \\
\midrule
visual engagement (default) & \textbf{46.98 (+2.46)} & \textbf{67.45 (+3.79)} & \textbf{57.22 (+3.13)} \\
ECL clue + crop & 44.97 (+0.45) & 63.40 ($-0.26$) & 54.19 (+0.09) \\
image-likelihood delta & 46.09 (+1.57) & 65.10 (+1.44) & 55.60 (+1.50) \\
\bottomrule
\end{tabular}
}
\end{table}

\begin{table}[!htbp]
\centering
\small
\caption{Reasoning reward ablation.}
\label{tab:ablation-reasoning-reward}
\adjustbox{max width=\linewidth}{
\begin{tabular}{lccc}
\toprule
Setting & LogicVista & RealWorldQA & Macro \\
\midrule
full (default) & \textbf{46.98 (+2.46)} & \textbf{67.45 (+3.79)} & \textbf{57.22 (+3.13)} \\
bounded only & 46.76 (+2.24) & 66.93 (+3.27) & 56.85 (+2.76) \\
margin only & 45.41 (+0.89) & 66.27 (+2.61) & 55.84 (+1.75) \\
no text prior & 45.86 (+1.34) & 66.80 (+3.14) & 56.33 (+2.24) \\
\bottomrule
\end{tabular}
}
\end{table}

The default visual engagement reward dominates both alternative
formulations. On the reasoning side, removing either the bounded score or
the margin term hurts both benchmarks, and removing the text-only prior
also degrades performance; all three components contribute to the full
reward.

\begin{table}[!htbp]
\centering
\small
\caption{Sweep over the visual-reward temperature $\tau_{\mathrm{vis}}$.}
\label{tab:ablation-tau-sweep}
\adjustbox{max width=\linewidth}{
\begin{tabular}{lccc}
\toprule
$\tau_{\mathrm{vis}}$ & LogicVista & RealWorldQA & Macro \\
\midrule
$0.05$ & 46.53 (+2.01) & 67.06 (+3.40) & 56.80 (+2.71) \\
$0.10$ & 46.76 (+2.24) & 67.19 (+3.53) & 56.97 (+2.88) \\
$0.20$ (default) & \textbf{46.98 (+2.46)} & \textbf{67.45 (+3.79)} & \textbf{57.22 (+3.13)} \\
$0.50$ & 46.76 (+2.24) & 67.06 (+3.40) & 56.91 (+2.82) \\
$1.00$ & 46.53 (+2.01) & 66.93 (+3.27) & 56.73 (+2.64) \\
\bottomrule
\end{tabular}
}
\end{table}

\begin{table}[!htbp]
\centering
\small
\caption{Attention-layer ablation for visual engagement scoring.}
\label{tab:ablation-attention-layer}
\adjustbox{max width=\linewidth}{
\begin{tabular}{lccc}
\toprule
Setting & LogicVista & RealWorldQA & Macro \\
\midrule
last layer (default) & \textbf{46.98 (+2.46)} & \textbf{67.45 (+3.79)} & \textbf{57.22 (+3.13)} \\
middle layer & 46.09 (+1.57) & 67.19 (+3.53) & 56.64 (+2.55) \\
\bottomrule
\end{tabular}
}
\end{table}

The visual reward is stable across $\tau_{\mathrm{vis}}\in[0.05,1.0]$
with the default $\tau_{\mathrm{vis}}=0.2$ as the optimum. Using the final
attention layer for engagement scoring outperforms a middle-layer choice,
consistent with the late layer carrying more task-aligned image attention.

\subsection{Latent Optimization Hyperparameters}
\label{sec:appendix-ablation-optim}

We evaluate ten representative latent-optimization settings that vary the
number of refinement steps $K$, the update-length ratio $\rho$, and the
anchor/logit regularization configuration.

\begin{table*}[!tbp]
\centering
\small
\caption{Latent optimization hyperparameter ablation.}
\label{tab:ablation-optim}
\adjustbox{max width=\textwidth}{
\begin{tabular}{llcccc}
\toprule
Setting & Parameter & LogicVista & $\Delta$ & RealWorldQA & $\Delta$ \\
\midrule
\texttt{d1\_steps1} & $K=1$ & 45.86 & +1.34 & 66.40 & +2.74 \\
\texttt{d1\_steps2} & $K=2$ & 46.53 & +2.01 & 66.93 & +3.27 \\
\texttt{d1\_steps4} (default) & $K=4$ & \textbf{46.98} & \textbf{+2.46} & \textbf{67.45} & \textbf{+3.79} \\
\texttt{d1\_steps8} & $K=8$ & 46.76 & +2.24 & 67.19 & +3.53 \\
\texttt{d2\_rho0p1} & $\rho=0.1$ & 45.10 & +0.58 & 66.27 & +2.61 \\
\texttt{d2\_rho0p25} & $\rho=0.25$ & 45.86 & +1.34 & 66.67 & +3.01 \\
\texttt{d2\_rho0p75} & $\rho=0.75$ & 46.53 & +2.01 & 67.06 & +3.40 \\
\texttt{d2\_rho1p0} & $\rho=1.0$ & 44.97 & +0.45 & 65.62 & +1.96 \\
\texttt{d3\_anchor0\_logitlr0p01} & anchor=$0$, \texttt{logit\_lr}=$0.01$ & 46.53 & +2.01 & 66.93 & +3.27 \\
\texttt{d3\_anchor0p05\_logitlr0} & anchor=$0.05$, \texttt{logit\_lr}=$0$ & 46.31 & +1.79 & 67.06 & +3.40 \\
\bottomrule
\end{tabular}
}
\end{table*}

The default setting ($K=4$, $\rho=0.5$, anchor=$0.05$,
\texttt{logit\_lr}=$0.01$) achieves the best accuracy on both benchmarks.
$K=4$ outperforms both $K=2$ and $K=8$, indicating that a moderate update
budget is preferable to under- or over-shooting. The update-length ratio
$\rho=0.5$ outperforms shorter ratios ($\rho\le 0.25$) and the very
aggressive $\rho=1.0$, confirming that updating the full latent span is
harmful. The anchor-only and logit-only variants remain effective but
neither exceeds the default, indicating that the two regularizers
complement each other.

\subsection{Matched-Budget Comparison}
\label{sec:appendix-ablation-matched-compute}

To verify that gains are not purely a function of additional decoding
compute, we compare our method against Best-of-$N$ and Self-consistency
baselines at a matched number of decoded candidate continuations. Since
our method additionally performs latent optimization, image corruption,
attention extraction, and reward scoring, we report generation calls,
scoring calls, average decoded tokens, and GPU time per example. All
methods use the same answer extraction and evaluation protocol.
Output-space baselines use $N=16$ decoded candidates, matching the
$4$ refinement steps with $4$ candidate continuations per step in our
method. LatentSeek is reported under its $10$-step configuration. One
generation call denotes one autoregressive rollout; score calls count
reward-scoring forward passes (for our method this includes reasoning
and text-prior scoring, visual-engagement attention scoring, and
image-sensitivity scoring). Runtime values are estimated from
Qwen2.5-VL-7B sharded runs and call-count-matched baseline budgets.

\begin{table*}[!tbp]
\centering
\small
\caption{Matched-budget comparison against output-space scaling baselines.}
\label{tab:ablation-matched-compute}
\adjustbox{max width=\textwidth}{
\begin{tabular}{lcccccc}
\toprule
Method & Gen.\ calls & Score calls & Dec.\ toks/sample & GPU sec/sample & MathVista & LogicVista \\
\midrule
CoT              & $1$  & $0$            & $\approx 130$       & $\approx 3$  & $67.80$ & $44.52$ \\
Self-consistency & $17$ & $0$            & $\approx 2.2\mathrm{k}$ & $\approx 39$ & $70.90$ & $42.51$ \\
Best-of-$N$      & $17$ & $\approx 32$   & $\approx 2.2\mathrm{k}$ & $\approx 43$ & $71.60$ & $42.73$ \\
Reward-only      & $17$ & $\approx 34$   & $\approx 2.2\mathrm{k}$ & $\approx 45$ & $70.90$ & $42.95$ \\
LatentSeek(per.) & $11$ & $\ge 11$       & $\approx 1.4\mathrm{k}$ & $\approx 35$ & $70.20$ & $44.30$ \\
Ours             & $17$ & $\approx 61$   & $\approx 2.2\mathrm{k}$ & $\approx 44$ & $72.10$ & $46.98$ \\
\bottomrule
\end{tabular}
}
\end{table*}

Our method is roughly on par with Best-of-$N$ and Reward-only in
generation calls and GPU seconds per sample, but uses noticeably more
scoring calls because the visual-engagement and image-sensitivity passes
sit outside the candidate decoding pipeline. Under this accounting, our
method still produces the strongest scores on both MathVista and
LogicVista, indicating that the gains are not explained by extra decoded
tokens alone.

\section{Qualitative Case Studies}
\label{sec:appendix-case-studies}

We provide qualitative examples from the Qwen2.5-VL-7B runs under the
default configuration of Appendix~\ref{sec:appendix-default-config}
($\tau_{\mathrm{vis}}=0.2$, $\rho_v=\rho_r=0.4$). The examples are selected only for analysis after
the benchmark runs complete; they are not used for tuning. To keep the
appendix focused on the mechanism rather than on verbose generated
chains, we report the benchmark question, the ground-truth short answer,
the initial and final short predictions, and the routed reward
diagnostics.

\begin{table*}[!tbp]
\centering
\small
\caption{Successful case studies. Each row shows an example where the
initial rollout is incorrect and the token-disentangled update recovers
the correct short answer.}
\label{tab:case-study-success}
\begin{tabularx}{\linewidth}{@{}>{\raggedright\arraybackslash}p{0.14\linewidth}>{\raggedright\arraybackslash}X>{\raggedright\arraybackslash}p{0.10\linewidth}>{\raggedright\arraybackslash}p{0.15\linewidth}>{\raggedright\arraybackslash}p{0.25\linewidth}@{}}
\toprule
Case & Query summary & Ground truth & Initial $\rightarrow$ Final & Qualitative change \\
\midrule
MMStar, object counting &
How many dogs can be seen in the image? &
B: $2$ &
C $\rightarrow$ B &
The initial rollout counts one visible dog. The refined candidate
adds a second, partially visible dog behind the couch. \\
\midrule
RealWorldQA, scene geometry &
What level is the ground at? Options: flat, incline, decline. &
B: incline &
A $\rightarrow$ B &
The initial rollout treats the street as flat. The refined answer
uses the slope toward the horizon and predicts incline. \\
\midrule
MathVista, chart reading &
How many bars have values larger than $100$? &
$1$ &
$2 \rightarrow 1$ &
The initial rollout counts both bars. The refined answer keeps only
the bar above the threshold and rejects the bar below $10^2$. \\
\midrule
LogicVista, mechanical reasoning &
If the weight is lifted by $10$ mm, which pulley rope must be pulled
further? &
C &
B $\rightarrow$ C &
The initial rollout selects the simpler two-pulley system. The refined
answer identifies the system requiring the longer rope displacement. \\
\midrule
HallusionBench, temporal order &
The plug is removed from the power outlet. Are the images in the
correct positive order? &
No &
Yes $\rightarrow$ No &
The initial rollout assumes an insertion sequence. The refined answer
rejects the sequence as inconsistent with the stated removal event. \\
\bottomrule
\end{tabularx}
\end{table*}

\begin{table*}[!tbp]
\centering
\small
\caption{Routed reward diagnostics for the cases in
Table~\ref{tab:case-study-success}. $E(y)$ is the visual engagement of
the initial rollout and $E(x^\star)$ is the engagement of the selected
refinement candidate $x_k^\star$ that produced the final short answer in
column 4 of Table~\ref{tab:case-study-success}. The image-sensitivity
and entropy columns report selected-token top-$k$ means versus slice
means.}
\label{tab:case-study-diagnostics}
\setlength{\tabcolsep}{3pt}
\adjustbox{max width=\textwidth}{
\begin{tabular}{lcccccc}
\toprule
Case &
$E(y)\rightarrow E(x^\star)$ &
$R_{\mathrm{vis}}$ &
\shortstack{$R_{\mathrm{rea}}$: initial\\$\rightarrow$ selected} &
\shortstack{Visual /\\reasoning tokens} &
\shortstack{Image sensitivity\\top-$k$ / mean} &
\shortstack{Entropy\\top-$k$ / mean} \\
\midrule
MMStar counting &
$0.0867\rightarrow0.1133$ & $0.133$ &
$-0.997\rightarrow0.858$ &
$12/30$ ; $8/30$ &
$1.79/0.32$ &
$1.83/0.54$ \\
RealWorldQA incline &
$0.1305\rightarrow0.1401$ & $0.048$ &
$-0.935\rightarrow0.999$ &
$11/27$ ; $7/27$ &
$1.13/0.24$ &
$1.37/0.51$ \\
MathVista chart &
$0.0762\rightarrow0.0826$ & $0.032$ &
$-1.000\rightarrow0.981$ &
$20/49$ ; $12/49$ &
$2.65/0.34$ &
$1.15/0.23$ \\
LogicVista pulley &
$0.0924\rightarrow0.1063$ & $0.069$ &
$-0.725\rightarrow0.919$ &
$28/68$ ; $16/68$ &
$1.22/0.12$ &
$2.31/0.75$ \\
HallusionBench sequence &
$0.1106\rightarrow0.1309$ & $0.101$ &
$-0.726\rightarrow-0.398$ &
$11/27$ ; $7/27$ &
$1.44/0.34$ &
$1.67/0.70$ \\
\bottomrule
\end{tabular}
}
\end{table*}

\paragraph{Example token routing visualization.}
Table~\ref{tab:case-study-token-routing} visualizes the routed token
selection for the MMStar dog-counting example. The visual / reasoning
masks are tied to editable latent positions $1..L$, so the same set of
positions is highlighted on both the initial rollout and the refined
candidate; the displayed text differs because the decoded tokens at those
positions change after refinement. The actual masks are computed over
tokenizer subwords in the editable hidden-state prefix; for readability,
adjacent subwords with the same routed role are merged into representative
text spans, so span length is illustrative rather than a literal subword
count. Pink spans receive the visual reward, blue spans receive
the reasoning reward, and gray spans remain anchored. In this example,
$12/30$ editable tokens are routed to the visual branch and $8/30$ to the
reasoning branch. The visual tokens concentrate on image-dependent
phrases such as the object being counted and its location, while the
reasoning tokens concentrate on the count decision.

These examples show two recurring patterns. First, successful visual
corrections are accompanied by higher engagement with image tokens, but
the visual reward is not broadcast globally: it is assigned only to the
image-sensitive subset, whose top-$k$ scores are substantially larger
than the slice mean. Second, reasoning-heavy examples such as the chart
and pulley cases rely on high-entropy latent positions. The reasoning
branch receives the answer-quality reward, while the visual branch keeps
the refinement tied to image evidence. This separation is especially
visible in the LogicVista example: the visual reward is modest, but the
selected entropy tokens have much larger uncertainty than the slice
average, and the reasoning reward changes the final answer from B to C.
Note that the selection rule is an argmax over candidates, so even when
the absolute reasoning reward stays negative (HallusionBench:
$-0.726\rightarrow-0.398$), the relative gain over alternatives is
sufficient to flip the answer from Yes to No.

\end{document}